\documentclass
[%
	draftclsnofoot,  
	peerreviewca,  
]%
{IEEEtran}%

\IEEEoverridecommandlockouts
\usepackage{cite}
\usepackage{amsmath,amssymb,amsfonts}
\usepackage{algorithmic}
\usepackage{graphicx}
\usepackage{textcomp}
\usepackage{xcolor}

\graphicspath{{figures/}{images/}{fig/}}  
\DeclareGraphicsExtensions{.pdf,.jpeg,.png}

\ifCLASSOPTIONcompsoc
	\usepackage
	[%
		caption=false,
		font=normalsize,
		labelfont=sf,
		textfont=sf,
	]{subfig}
\else
	\usepackage
	[%
		caption=false,
		font=footnotesize,
	]{subfig}
\fi

\usepackage{tikz}
\usetikzlibrary
{
	patterns,
	positioning,
	decorations.pathreplacing,
}

\usepackage[hidelinks]{hyperref}	
\usepackage{siunitx}				
\usepackage{xspace}					

\input{custommacros.sty}

\newcommand{\emergencyvehicle}{EV}
\newcommand{\emergencyvehicleformula}{\text{\emergencyvehicle}}
\newcommand{\emergencyvehicletext}{\textsc{\small \emergencyvehicle}\xspace}

\newcommand{\maximumthroughput}{\textsc{\small MT}\xspace}
\newcommand{\delaysensitive}{\textsc{\small DS}\xspace}
\newcommand{\maxminfair}{\textsc{\small MMF}\xspace}

\newcommand{\userindex}					{n}
\newcommand{\numusers}					{N}
\newcommand{\jobindex}					{j}

\newcommand{\jobsset}					{\mathbb{J}}
\newcommand{\resourceblock}				{u}
\newcommand{\numresourcesavailable}		{U}
\newcommand{\distanceuser}				{d}

\newcommand{\resourceblockrequested}{\resourceblock_{\jobindex,\,\text{req}}}
\newcommand{\resourceblockscheduled}{\resourceblock_{\userindex,\,\text{sx}}}
\newcommand{\resourceblocksizemaxuser}{\resourceblock_{\userindex,\,\text{max}}}
\newcommand{\jobsfailedset}{\jobsset_{\text{fail}}}
\newcommand{\jobsuserset}{\jobsset_{\userindex}}
\newcommand{\jobsnewset}{\jobsset_{\userindex,\,\text{new}}}
\newcommand{\numusersrequesting}{\numusers_{\text{req}}}

\newcommand{\timetotimeout}				{v}
\newcommand{\lowesttimetotimeout}		{l}
\newcommand{\packetrate}				{k}
\newcommand{\timeouts}					{m}
\newcommand{\timetotimeoutjob}{\timetotimeout_{\jobindex}}
\newcommand{\timetotimeoutinitialuser}{\timetotimeout_{\userindex,\,\text{init}}}
\newcommand{\lowesttimetotimeoutuser}{\lowesttimetotimeout_{\userindex}}

\newcommand{\rewardtimeout}{\reward_L}
\newcommand{\rewardtimeoutev}{\reward_{L,\, \emergencyvehicleformula}}
\newcommand{\rewardcapacity}{\reward_C}
\newcommand{\rewardpacketrate}{\reward_P}
\newcommand{\weightrewardtimeout}{\weight_L}
\newcommand{\weightrewardtimeoutev}{\weight_{L,\, \emergencyvehicleformula}}
\newcommand{\weightrewardcapacity}{\weight_C}
\newcommand{\weightrewardpacketrate}{\weight_P}

\newcommand{\probability}				{p}
\newcommand{\probabilityjobcreation}{\probability_{\jobindex}}

\newcommand{\actionindex}				{i}

\definecolor{ccolor1}{RGB}{208,27,136}
\definecolor{ccolor2}{RGB}{37,71,150}
\definecolor{ccolor3}{RGB}{48,123,59}
\definecolor{ccolor4}{RGB}{202,160,35}

\begin{document}

\title
{%
	Deep Reinforcement Model Selection\\for Communications Resource Allocation\\in On-Site Medical Care
	\thanks
	{%
		This work was partly funded by the German Ministry of Education and Research (BMBF) under grant 16KIS1028 (MOMENTUM).
		
		This work was accepted for presentation at IEEE WCNC2022.
	}%
}%

\author{%
	\IEEEauthorblockN{%
		Steffen~Gracla%
		,
		Edgar~Beck%
		,
		Carsten~Bockelmann
		and Armin~Dekorsy%
	}%
	\IEEEauthorblockA{%
		Dept. of Communications Engineering, University of Bremen, Bremen, Germany\\
		{Email: \{gracla, beck, bockelmann, dekorsy\}@ant.uni-bremen.de}
	}%
}%
\maketitle%
%



\begin{abstract}
	Greater capabilities of mobile communications technology enable the interconnection of on-site medical care at a scale previously unavailable.
	However, embedding such critical, demanding tasks into the already complex infrastructure of mobile communications has proven challenging.
	This paper explores a resource allocation scenario where a scheduler must balance mixed performance metrics among connected users.
	To fulfill this resource allocation task, we present a scheduler that adaptively switches between different model-based scheduling algorithms.
	We make use of a deep Q-Network~(\dqn) to learn the benefit of selecting a scheduling paradigm for a given situation, combining advantages from model-driven and data-driven approaches.
	The resulting ensemble scheduler is able to combine its constituent algorithms to maximize a sum-utility cost function while ensuring performance on designated high-priority users.
\end{abstract}


\section{Introduction}\label{sec:intro}
The optimal allocation of communication resources is part of the perpetual race for better performance in mobile communications.
As new technologies such as high resolution video streaming and vehicular communications emerge, the demands to be fulfilled by a scheduler are becoming increasingly heterogeneous, resulting in complex optimization tasks that must be solved in real time.

In recent years, deep learning~(\deeplearning) methods have distinguished themselves with success in complex tasks, including image classification~\cite{krizhevsky2012imagenet} and control~\cite{mnih2015human}, among others.
\deeplearning, as a subset of machine learning~(\machinelearning), follows a different paradigm compared to classic model-based approaches. Instead of designing an optimal algorithm for a given goal, an algorithm is iteratively approximated based on training data. Without the need to explicitly model the underlying processes, the issue of modeling complexity is sidestepped.
This characteristic has sparked an interest in research that applies \deeplearning to the highly performance-driven field of communication systems.

In particular, increasing the performance and capabilities of mobile communication systems creates new prospects in emergency patient care.
For example, video, vitals, or specialist input can be exchanged wirelessly between on-site medical professionals and hospital staff, yielding a significant head start in time-sensitive treatments~\cite{zanatta2016pre}.
Nevertheless, a deliberate approach is required due to the low tolerance for delays and outliers in medical tasks.

Various forays have been made into integrating \deeplearning methods with resource allocation tasks, mostly based on deep Q-Networks~(\dqn)~\cite{joseph2019towards, ye2019deep, al2020learn} and actor-critic methods~\cite{huang2020deep}.
However, model-based approaches have long thrived, thanks to the valuable expert knowledge that shapes their design. Ideally, incorporating this expert knowledge into data-based methods could govern and stabilize the learning process and produce more tractable algorithms~\cite{shlezinger2020model}.

Therefore, we investigate a combined approach that offers some of the advantages of both \deeplearning and model-based approaches.
For the task of scheduling discrete resources in vehicle-to-base-station communication, we implement a group of simple, model-based scheduling algorithms.
We then train a \dqn to select the best model-based algorithm in a given situation, optimizing the long term effect on typical performance indicators: time-outs, sum capacity, and packet rate.
Particular attention is paid to the performance of Emergency Vehicles~(\emergencyvehicletext) as a priority class of users.
In this way, we are able to use data-driven \deeplearning to find an approximately optimal solution to a complex scheduling problem while still using explicitly modeled algorithms.

\section{Setup \& Notation}
In the following, we introduce our system model. We describe the communication channel between a user and the base station, the resources available for allocation, and the generation of jobs. We then formulate the resource allocation problem. After these technical specifications, we move on to the \dqn scheduler design.

\subsection{Resource Allocation System Model}
In our system model, a number of \( \numusers \)~user vehicles are connected to a base station, moving on a 2D plane in a grid akin to the Manhattan-model of movement~\cite{aschenbruck2008survey}.
At any discrete time instance~\( \timeindex \), vehicles~\( \userindex \) will take a fixed-size step in a direction of movement that is selected randomly. In doing so, the vehicles have a \SI{98}{\percent} chance of re-selecting their prior direction of movement, \SI{0}{\percent} chance of selecting the direction opposite to their prior movement, and uniform chance of selecting a \ang{90} turn left, right, or stopping.

The communication channel between the base station and a user vehicle~\( \userindex \) is characterized by its power gain
\begin{align}
	\channelquality_\userindex[\timeindex]
	=
	| \tilde{\channelquality}_{\userindex}[\timeindex] |^{\num{2}} \cdot \pathloss_{\userindex}[\timeindex].
\end{align}
The power gain is known to the base station. For each simulation time step~\( \timeindex \), fading channel amplitudes~\( | \tilde{\channelquality}_{\userindex}[\timeindex] | \) are randomly selected from a {Rayleigh distribution} and multiplied with a distance-proportional path loss factor
\begin{align}
	\pathloss_\userindex[\timeindex]
	=
	\min\left(
	1,\,\left(\distanceuser_\userindex[\timeindex]\right)^{\num{-1}}
	\right),
\end{align}
where \( \distanceuser_{\userindex}[\timeindex] \)~is a vehicle's distance from the base station.
The path loss factor ensures a degree of correlation of the power gain~\( \channelquality_{\userindex}[\timeindex] \) over time.
While exponents of~\( \num{-2} \) or lower are more typically assumed in modeling real-life path loss, we selected an exponent of~\( \num{-1} \) to reduce the spread of the path loss values encountered. For the purpose of this paper, this lessens the computation required in the subsequent learning task.

\begin{figure}[tb]
	\centering
	\begin{tikzpicture}
\tikzstyle{rb} = [rectangle, draw, minimum width=.3cm, minimum height=.3cm]
\tikzstyle{rbu1} = [rb, pattern=north west lines, pattern color=ccolor2]
\tikzstyle{rbu2} = [rb, pattern=crosshatch dots, pattern color=ccolor1]
\tikzstyle{rbu3} = [rb, pattern=grid, pattern color=ccolor4]

\newcommand{\xanchor}{-1.0}
\newcommand{\xstep}{0.4}
\newcommand{\ystep}{0.4}

\node (r11) at (\xanchor + 0 * \xstep, +2.0 - 0 * \ystep) [rbu1] {};
\node (r12) at (\xanchor + 0 * \xstep, +2.0 - 1 * \ystep) [rbu2] {};
\node (r13) at (\xanchor + 0 * \xstep, +2.0 - 2 * \ystep) [rbu2] {};
\node (r14) at (\xanchor + 0 * \xstep, +2.0 - 3 * \ystep) [rbu2] {};
\node (r15) at (\xanchor + 0 * \xstep, +2.0 - 4 * \ystep) [rbu2] {};

\node(r21) [rb] at (\xanchor + 1 * \xstep, +2.0 - 0 * \ystep) {};
\node(r22) [rb] at (\xanchor + 1 * \xstep, +2.0 - 1 * \ystep) {};
\node(r23) [rb] at (\xanchor + 1 * \xstep, +2.0 - 2 * \ystep) {};
\node(r24) [rb] at (\xanchor + 1 * \xstep, +2.0 - 3 * \ystep) {};
\node(r25) [rb] at (\xanchor + 1 * \xstep, +2.0 - 4 * \ystep) {};

\node(r31) [rb] at (\xanchor + 2 * \xstep, +2.0 - 0 * \ystep) {};
\node(r32) [rb] at (\xanchor + 2 * \xstep, +2.0 - 1 * \ystep) {};
\node(r33) [rb] at (\xanchor + 2 * \xstep, +2.0 - 2 * \ystep) {};
\node(r34) [rb] at (\xanchor + 2 * \xstep, +2.0 - 3 * \ystep) {};
\node(r35) [rb] at (\xanchor + 2 * \xstep, +2.0 - 4 * \ystep) {};

\node(r41) [rb] at (\xanchor + 3 * \xstep, +2.0 - 0 * \ystep) {};
\node(r42) [rb] at (\xanchor + 3 * \xstep, +2.0 - 1 * \ystep) {};
\node(r43) [rb] at (\xanchor + 3 * \xstep, +2.0 - 2 * \ystep) {};
\node(r44) [rb] at (\xanchor + 3 * \xstep, +2.0 - 3 * \ystep) {};
\node(r45) [rb] at (\xanchor + 3 * \xstep, +2.0 - 4 * \ystep) {};

\node(r51) [rb] at (\xanchor + 4 * \xstep, +2.0 - 0 * \ystep) {};
\node(r52) [rb] at (\xanchor + 4 * \xstep, +2.0 - 1 * \ystep) {};
\node(r53) [rb] at (\xanchor + 4 * \xstep, +2.0 - 2 * \ystep) {};
\node(r54) [rb] at (\xanchor + 4 * \xstep, +2.0 - 3 * \ystep) {};
\node(r55) [rb] at (\xanchor + 4 * \xstep, +2.0 - 4 * \ystep) {};

\node(job111) [rbu1] at (-3.0, +2.2) {};
\node(job112) [rbu1] at (-2.6, +2.2) {};
\node(job113) [rbu1] at (-2.2, +2.2) {};
\node(job114) [rbu1] at (-3.0, +1.8) {};
\node(job115) [rbu1] at (-2.6, +1.8) {};

\node(job211) [rbu2] at (-3.0, +1.3) {};
\node(job212) [rbu2] at (-2.6, +1.3) {};
\node(job213) [rbu2] at (-2.2, +1.3) {};

\node(job221) [rbu2] at (-4.6, +1.3) {};
\node(job222) [rbu2] at (-4.2, +1.3) {};
\node(job223) [rbu2] at (-3.8, +1.3) {};
\node(job224) [rbu2] at (-4.6, +0.9) {};
\node(job225) [rbu2] at (-4.2, +0.9) {};
\node(job226) [rbu2] at (-3.8, +0.9) {};

\node(job311) [rbu3] at (-3.0, +0.4) {};
\node(job312) [rbu3] at (-2.6, +0.4) {};

\draw[]
	(\xanchor + 5 * \xstep, 0.2)
	-- node
		[align=center, right]
		{$\numresourcesavailable$}
	(\xanchor + 5 * \xstep, 2.2);
	
\draw [-{stealth}]
	(\xanchor - 0.2, 0.0)
	-- node
		[below, align=left]
		{Time $\timeindex$}
	(\xanchor - 0.2 + 5 * \xstep, 0.0);
	
\draw[-]
	(-4.8, 0.1)
	--
	(-4.8, -0.1)
	-- node
		[below, align=center]
		{Job Queue}
	(-2.0, -0.1)
	--
	(-2.0, 0.1);
	
\draw [-{stealth}]
	(-1.8, 1.2)
	--
	(\xanchor - 0.3, 1.2);

\node(user1) [rbu1] at (-7.0, +2.2) {};
\node(user2) [rbu2] at (-7.0, +1.7) {};
\node(user3) [rbu3] at (-7.0, +1.2) {};
\node(user1t) [] at (-6.1, +2.2) {User 1};
\node(user2t) [] at (-6.1, +1.7) {User 2};
\node(user3t) [] at (-6.1, +1.2) {User 3};

\draw [-] (-4.5, 1.5) -- +(-0.1, 0.2) node[above] {\( \resourceblockrequested=\num{6} \)};

\end{tikzpicture}
	\caption
	{%
		Jobs consisting of a number of discrete resource blocks arrive in the job queue.
		A scheduler is tasked with assigning them to a limited number~\( \numresourcesavailable \) of resources.
	}%
	\label{fig:resourceblocks}
\end{figure}
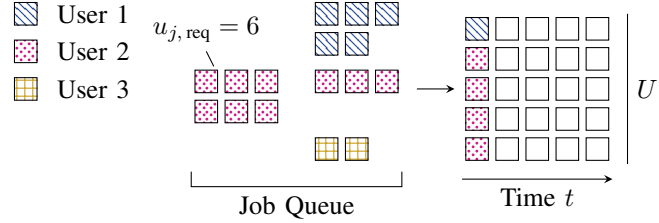

The base station has access to a limited number~\( \numresourcesavailable \) of discrete resource blocks available for allocation, as shown in \reffig{fig:resourceblocks}.
These resources can be filled with jobs~\( \jobindex \) from the job queue.
At each time step~\( \timeindex \), new jobs are generated at a probability~\( \probabilityjobcreation \) per user.
A job~\( \jobindex \) is assigned to a specific user~\( \userindex \) and is defined by two attributes, a request size~\( \resourceblockrequested[\timeindex] \) in discrete resource blocks, and a time-to-timeout~\( \timetotimeoutjob[\timeindex] \) in discrete simulation steps.
We define the set~\( \jobsset[\timeindex] \) of all jobs in the job queue in time step~\( \timeindex \) and the subset~\( \jobsuserset[\timeindex] \) of the jobs in this queue assigned to user~\( \userindex \) in~\(\timeindex \).

The job attributes' initial values are governed by a user profile that declares a maximum job size~\( \resourceblocksizemaxuser \) and initial time-to-timeout~\( \timetotimeoutinitialuser \) for users~\( \userindex \) of that profile.
Upon generation, the job is assigned a size~\( \resourceblockrequested[\timeindex] \) selected from a discrete uniform distribution, \( 
	{\resourceblockrequested[\timeindex] \sim \mathbb{U}\{1,\, \resourceblocksizemaxuser\}} 
\). When a number of the base station's resources~\( \numresourcesavailable \) is allocated to a job~\( \jobindex \), that job's remaining size is decreased accordingly, while a lifetime count~\( \resourceblockscheduled[\timeindex] \) of resources scheduled to user~\( \userindex \) is increased.
The jobs' initial time-to-timeout~\( {
	\timetotimeoutjob[\timeindex] \gets \timetotimeoutinitialuser
} \) is decremented for each time step~\( \timeindex \) that passes without the job being fully scheduled. Once the time-to-timeout~\( \timetotimeoutjob[\timeindex] \) reaches zero, the job is discarded from the job queue and added to a set~\( \jobsfailedset[\timeindex] \) for that time step~\( \timeindex \), for use in performance metric calculation.

\subsection{Problem Statement}
A scheduler is tasked with assigning the limited number~\( \numresourcesavailable \) of resource blocks to the jobs in queue.
Three metrics are selected to gauge the performance of a scheduling algorithm:
\begin{enumerate}
	\item \( \rewardtimeout[\timeindex] \): Resource blocks discarded from timeout
	\item \( \rewardpacketrate[\timeindex] \): Global packet rate achieved
	\item \( \rewardcapacity[\timeindex] \): Global channel capacity achieved
\end{enumerate}

Firstly, the global \emph{sum of resource blocks discarded} due to timing out
\begin{align}
	\rewardtimeout[\timeindex]
	=
	\sum_{\jobindex \in \jobsfailedset[\timeindex]} \resourceblockrequested[\timeindex] 
\end{align}
should be minimized.

Secondly, a packet rate is introduced as the lifetime ratio of resources requested and scheduled for a user~\( \userindex \).
We define the set~\( \jobsnewset[\timeindex] \) as the set of new jobs assigned to user~\( \userindex \), generated in time step~\( \timeindex \). Using the lifetime sum of discrete resources~\( \resourceblockscheduled[\timeindex] \) scheduled to the jobs of user~\( \userindex \), we calculate the \emph{sum packet rate} over all users:
\begin{align}
	\rewardpacketrate[\timeindex]
	= 	\sum_{\userindex=\num{1}}^{\numusers}
		\frac{
			\resourceblockscheduled[\timeindex]
		}{
			\sum_{\tilde{\timeindex}=\num{1}}^{\timeindex}
			\sum_{\jobindex \in \jobsnewset\left[\tilde{\timeindex}\right]}
			\resourceblockrequested\left[\tilde{\timeindex}\right]
		}
	= 	\sum_{\userindex=\num{1}}^{\numusers}
		\packetrate_{\userindex}[\timeindex]
	.
\end{align}
A strong sum packet rate performance is achieved by a scheduler that does not neglect any single user.

Last, the \emph{sum rate capacity} achieved by each transmission is calculated using the Signal-to-Noise ratio~(\SNR) resulting from the selected channels instantaneous fading characteristics as
\begin{align}
	\rewardcapacity[\timeindex]
	=
	\sum_{\userindex=1}^{\numusers}
	\log
	\left(
		1 +
		\channelquality_{\userindex}[\timeindex]
		\frac{\signalpower}{\noisepower}
	\right) 
	=
	\sum_{\userindex=1}^{\numusers}
	\log
	\left(
		1 + \SNR_{\userindex}[\timeindex]
	\right) 
\end{align}
for a Gaussian input alphabet, where signal power~\( \signalpower \) and expected noise power~\( \noisepower \) are fixed for all vehicles~\( \userindex \).

The scheduler is tasked with balancing all performance metrics, thus, we collect all target metrics in a weighted sum utility
\begin{align}
	\tilde{\reward}[\timeindex] = 
	  \weightrewardcapacity\rewardcapacity[\timeindex]
	+ \weightrewardpacketrate \rewardpacketrate[\timeindex]
	- \weightrewardtimeout \rewardtimeout[\timeindex],
\end{align}
with respective tunable weights~\( \weightrewardcapacity, \weightrewardpacketrate, \weightrewardtimeout \). Additionally, we designate a priority class of \emergencyvehicletext-type users. Their significance is communicated to the optimization process by adding \emergencyvehicletext timeouts~\( \rewardtimeoutev[\timeindex] \) to the weighted sum utility with their own tunable weight~\( \weightrewardtimeoutev \):
\begin{align}\label{eq:reward}
	\reward[\timeindex]
	&=	  \tilde{\reward}[\timeindex]
		- \weightrewardtimeoutev \rewardtimeoutev[\timeindex]\nonumber\\
	&=	  \weightrewardcapacity \rewardcapacity[\timeindex]
		+ \weightrewardpacketrate \rewardpacketrate[\timeindex]
		- \weightrewardtimeout \rewardtimeout[\timeindex]
		- \weightrewardtimeoutev \rewardtimeoutev[\timeindex].
\end{align}

\section{Algorithm Selection Approach}\label{sec:deep}
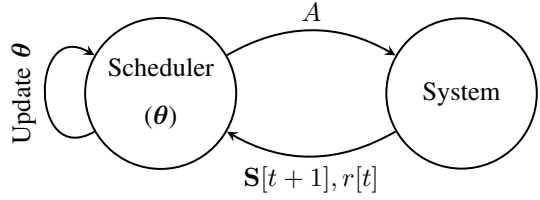
\begin{figure}[tb]
	\centering
	\tikzset{every picture/.style={line width=0.75pt}} 

\begin{tikzpicture}
\tikzstyle{A} = [->, >=stealth]

\node (scheduler)
	at (0, 0)
	[draw, circle, align=center, minimum width=2cm]
	{Scheduler \\ ($\parameters$)};

\node (system)
	at (4, 0)
	[draw, circle, align=center, minimum width=2cm]
	{System};

\draw[A]
	(scheduler) to [out=30, in=150] node[above]{$\actionsca$} (system);
	
\draw[A]
	(system) to [out=210, in=-30] node [below]{$\statevec[\timeindex+1], \reward[\timeindex]$} (scheduler);

\draw[A]
	(scheduler)
		to [out=210, in=150, min distance=10mm]
		node [rotate=90, above, align=center] {Update $\parameters$}
	(scheduler);

\end{tikzpicture}
	\caption
	{%
		In the desired RL loop, the parametrized scheduler interacts with the system by selecting an action \( \actionsca \) and receiving a resulting reward~\( \reward[\timeindex] \) and new system state~\( \statevec[\timeindex+1] \).
		Based on these experiences, the scheduler updates its parametrization~\( \parameters \) to promote high-reward actions and demote low reward actions.
	}%
	\label{fig:RL_Training}
\end{figure}

Where model-based approaches struggle to find an optimal solution, deep Reinforcement Learning~(\reinforcementlearning) may be applied to learn a function that approximates the optimal algorithm along the domain of reasonable input data.
\reffig{fig:RL_Training} schematically depicts the desired learning process.
At the same time, some applications require boundary conditions to be met, which tend to be easy to formulate in model-based algorithms but cannot be enforced directly in standard \reinforcementlearning approaches.

Our scheduler learns to adaptively switch between a selection of model-based algorithms to tackle this problem.
In order to select the best algorithm, the scheduler makes use of a deep Q-Network~(\dqn)~\cite{sutton2018reinforcement} to learn the long-term benefit of selecting each operation mode in a given situation.
As a result, model-based design benefits, such as hard performance guarantees and human interpretability, are provided by the pool of model-based algorithms.
Meanwhile, the superposition of algorithms allows for greater performance than each individual algorithm on flexible goal metrics.
As an additional benefit, the individual model-based algorithms do not have to be sophisticated enough to perform well in every circumstance, so long as the overall selection is rich enough to serve any problem.

\subsection{Model-Based Algorithms}
Four model-based scheduling algorithms are implemented for the \dqn to select from:~\cite{schmidt2018analyse}

(1) A \textbf{Maximum Throughput\,(\maximumthroughput)} scheduler that allocates as many resources as requested by order of descending channel power gains.

(2) A \textbf{Max-Min-Fair\,(\maxminfair)} scheduler looks to distribute the available resources~\( \numresourcesavailable \) performantly but fairly among the number~\( \numusersrequesting[\timeindex] \) of users that have jobs assigned to them in the job queue in time step~\( \timeindex \). It allocates by order of priority
\( 
	{\channelquality_{\userindex}[\timeindex]\,
	/\,
	\sum_{\jobindex \in \jobsuserset[\timeindex]}
	\resourceblockrequested[\timeindex]}
 \), favoring good channels and small requests, but allocates at most an equal share
\( 
	\left\lfloor
		\numresourcesavailable
		/
		\numusers_{\text{req}}[\timeindex]
	\right\rfloor
\).

(3) A \textbf{Delay Sensitive\,(\delaysensitive)} scheduler assigns a channel priority 
\begin{align*}
	{p}_{c, \userindex}[\timeindex]
	=
	\frac{
		\packetrate_{\userindex}[\timeindex]
	}
	{
		\sum_{q=1}^{\numusers} \packetrate_{q}[\timeindex]
	}
	\frac{
		\channelquality_{\userindex}[\timeindex]
	}
	{
		\sum_{q=1}^{\numusers} \channelquality_{q}[\timeindex]
	}
\end{align*}
given each users relative channel power gain~\( \channelquality_{\userindex}[\timeindex] \) and packet rate~\( \packetrate_{\userindex}[\timeindex] \).
The \delaysensitive scheduler further draws on the users sum timeouts~\(
	{\timeouts_{\userindex}[\timeindex] = \sum_{\tilde{\timeindex}=\num{1}}^{\timeindex} \reward_{L, \userindex}\left[\tilde{\timeindex}\right]} 
\) and lowest remaining time~\(
	{\lowesttimetotimeoutuser[\timeindex] = \min_{\jobindex \in \jobsuserset[\timeindex]} \timetotimeout_{\jobindex}[\timeindex]}
\) for a timeout urgency
\begin{align*}
	{p}_{l, \userindex}[\timeindex]
	=
	\frac{
		\timeouts_{\userindex}[\timeindex]
		/
		\lowesttimetotimeoutuser[\timeindex]
	}
	{
		\sum_{q=1}^{\numusers}
		\timeouts_{q}[\timeindex]
		/
		\lowesttimetotimeout_{q}[\timeindex]
	}
	.
\end{align*} 
Each user is allotted a share of the available resources according to the normalized, weighted priority vector
\begin{align*}
	\mathbf{p}[\timeindex]
	=
	(
	w_1\mathbf{p}_{c}[\timeindex]
	+
	w_2\mathbf{p}_{l}[\timeindex]
	)
	/
	(
	w_1
	+
	w_2
	)
	.
\end{align*}
Uniquely, this scheduler skips jobs that are about to time out if the allotted discrete resources are not sufficient to complete the job. In this case, the resources are freed for the next highest priority.

(4) An \textbf{\emergencyvehicletext Priority} scheduler assigns as many resources as requested to any \emergencyvehicletext{}s and distributes remaining resources one-by-one, randomly assigning them to requesting users.

\subsection{Deep Algorithm Selection}
Our deep learning scheduler consists of multiple elements: A pre- and post-processor, a \dqn, an~\( \argmax \) module, a memory module, and a learning algorithm that tunes the \dqn. In this section, we will introduce these modules and illustrate their interconnection.

First, to be able to make informed decisions, a small pre-processor prepares information about the current state of job queue and communication link as a system state vector~\( \statevec[\timeindex] \). Per user~\( \userindex \), the pre-processor summarizes
\begin{itemize}
	\item current queue length \(
		\statesca_{\userindex, 1}[\timeindex]
			= \sum_{\jobindex \in \jobsuserset[\timeindex]} \resourceblockrequested[\timeindex]
		\),
	\item channel power gains \(
		\statesca_{\userindex, 2}[\timeindex]
			= \channelquality_\userindex[\timeindex]
		\),
	\item average remaining time \(
		\statesca_{\userindex, 3}[\timeindex]
			= \frac{1}{\left| \jobsuserset[\timeindex] \right|} \sum_{\jobindex \in \jobsuserset[\timeindex]} \timetotimeout_{\jobindex}[\timeindex]
		\),
	\item minimum remaining time \(
		\statesca_{\userindex, 4}[\timeindex]
			= \min_{\jobindex \in \jobsuserset[\timeindex]} \timetotimeout_{\jobindex}[\timeindex]
		\),
	\item and past packet rate \(
		\statesca_{\userindex, 5}[\timeindex]
			= \packetrate_\userindex[\timeindex]
		\),
\end{itemize}
for a length of \( 5\numusers \) features for \( \numusers \) users.

For a given system state~\( \statevec[\timeindex] \) and choice of model-based algorithm~\( \actionsca_{\actionindex} \), we define the long term expected rewards
\begin{align}
	\label{eq:def_q}
	\longrewardsca(\statevec[\timeindex], \actionsca_{\actionindex})
	=
	\expectation
	\left[
	\sum_{z=\timeindex}^{\infty}
	\rewarddiscount^{z-\timeindex}
	\reward[\timeindex]
	\;\middle|\;
	\statevec = \statevec[\timeindex],\,
	\actionsca = \actionsca_{\actionindex}
	\right],
\end{align}
with reward~\( \reward \) as defined in~\refeq{eq:reward}. The long term rewards are discounted by an exponential factor~\(
	{\num{0} \leq \rewarddiscount \leq \num{1}}
\).
As depicted in \reffig{fig:dqn}, a \dqn is set up to output an estimate~\( \longrewardestimsca \) of the long term expected rewards~\( \longrewardsca \) for each choice of model-based algorithm~\( \actionsca_{\actionindex} \).
Given a perfect approximation~\(
	{\longrewardestimsca = \longrewardsca}
\), we maximize the long term expected rewards by simply selecting whichever action~\(
	\argmax_{\actionindex} \longrewardestimsca(\statevec[\timeindex],\,\actionsca_{\actionindex})
\) has the highest \emph{estimated} long term rewards.
Therefore, the learning goal is to update the \dqn parameters~\( \parameters \) such that the estimate is approximately close to the true long term reward~\( \longrewardsca \) for all possible~\(
	{(\statevec, \actionsca_{\actionindex})}
\), relying on the universal approximation property of neural networks~\cite{sutton2018reinforcement}.

\begin{figure}[tb]
	\centering
	\tikzset{every picture/.style={line width=0.75pt}} 

\begin{tikzpicture}
\tikzstyle{A} = [->, >=stealth]

\node(St)
	{
		$\statevec[\timeindex]$
	};

\node(dqn)
	[
		rectangle, rounded corners, draw, minimum width=2.5cm, minimum height=1cm,
		right = of St
	]
	{
		DQN ($\parameters$)
	};

\draw[A] (St) -- (dqn);

\node(q2)
	[right = of dqn]
	{$\longrewardestimsca(\statevec[\timeindex], \actionsca_2)$};
	
\node(q1)
	[above = .05cm of q2]
	{$\longrewardestimsca(\statevec[\timeindex], \actionsca_1)$};
	
\node(q3)
	[below = .05cm of q2]
	{$\longrewardestimsca(\statevec[\timeindex], \actionsca_3)$};

\draw[A]
	(dqn.east) to (q1.west);
\draw[A]
	(dqn.east) -- (q2);
\draw[A]
	(dqn.east) to (q3.west);

\draw
	[decorate, decoration={brace}]
	(q1.east) + (0.0, 0.2) -- (q1.east) -- node [rotate=90, below, align=center] {$\displaystyle\argmax_\actionsca$} (q3.east) -- +(0.0, 0.2);

\end{tikzpicture}
	\caption
	{%
		The DQN module estimates expected long term rewards~\(
		{\longrewardestimsca(\statevec[\timeindex], \actionsca_\actionindex)}
		\) for a given state~\( \statevec[\timeindex] \) and all available actions~\( \actionsca_{\actionindex} \), according to its current parametrization~\( \parameters \).
		In this case, actions~\( \actionsca_{\actionindex} \) are the available model-based algorithms. The scheduler then selects the algorithm~\( \actionsca_{\actionindex} \) with the highest expected long term reward.
	}%
	\label{fig:dqn}
\end{figure}

First, the scheduler must make experiences to learn from.
Using an \( \explorationchance \)-greedy exploration scheme, the scheduler initially explores the simulation by taking random actions, \ie selecting a random model-based algorithm for a given state.
States, decisions, and their direct outcome are recorded and stored in a replay buffer in the form of tuples
\begin{align}
	\experience
	=
	\left(
		\statevec[\timeindex],\,
		\actionsca[\timeindex],\,
		\reward[\timeindex],\,
		\statevec[\timeindex+1]
	\right)
	.
\end{align}

We highlight that an experience only contains the immediate reward~\( \reward[\timeindex] \), not the desired long-term rewards~\(
	{\longrewardsca(\statevec[\timeindex], \actionsca[\timeindex])}
\), \ie~\refeq{eq:def_q}.
However, we can extract additional information from~\( \statevec[\timeindex+1] \), the state that following the action. By again using our \dqn, we estimate the rewards following~\( \statevec[\timeindex+1] \) to construct a learning target
\begin{align}
	\label{eq:bootstrap}
	\longrewardestimsca_{\text{target}}(\experience)
	=
	\reward[\timeindex]
	+
	\rewarddiscount
	\max_\actionindex
		\longrewardestimsca(\statevec[\timeindex+1], \actionsca_\actionindex)
	,
\end{align}
that incorporates the information of~\( \reward[\timeindex] \) and ~\( \statevec[\timeindex + \num{1}] \) from the experience.
Using this target~\( {\longrewardestimsca_{\text{target}}(\experience)} \), an estimation error~\cite{sutton2018reinforcement}
\begin{align}
	\tderror
	=
	\longrewardestimsca_{\text{target}}\left( \experience \right)
	-
	\longrewardestimsca \left( \experience \right)
\end{align}
given the networks current parametrization can be calculated for any experience tuple from the buffer.

Following the principle of stochastic gradient descent~(\stochasticgradientdescent), the networks parameters~\( \parameters \) are then adjusted by sampling a mini batch of~\( \batchsize \) experiences from the buffer to minimize a mean square error cost
\begin{align}
	\cost
	=
	\frac{
		1
	}
	{
		\batchsize
	}
	\sum_{\batchindex=1}^{\batchsize}
	(
		\tderror_{\batchindex}
	)^2
\end{align}
for the batch.
Minibatch parameter updates are carried out every time a new experience is made.

As training progresses and the schedulers understanding of the simulation environment improves, the probability~\( \explorationchance \) of selecting random exploration actions is gradually decreased, relying more and more on the network to make decisions. When prompted, at the beginning of a time step~\( \timeindex \), the network will estimate the expected long term rewards~\( \longrewardestimsca \) for selecting any of the available model-based algorithms~\( \actionsca_{\actionindex} \), given the current state~\( \statevec[\timeindex] \) and the networks current parameters~\( \parameters \). The scheduler then selects the model-based algorithm~\( \actionsca_{\actionindex} \) with the highest expected long term rewards.

To increase training efficiency, we implement optimizations to the base \dqn learning method: 
\begin{itemize}
	\item An infrequently updated network copy is used for the bootstrapping in \refeq{eq:bootstrap} to increase training stability (\emph{Target Network},~\cite{mnih2015human})
	\item Sampling of experiences from the buffer is weighted proportional to the experiences' estimation error magnitude (\emph{Prioritized Replay},~\cite{hessel2018rainbow})
	\item The neural network structure is altered to seperately learn the contribution of state and action to the reward estimate (\emph{DuelingDQN},~\cite{hessel2018rainbow})
\end{itemize}

\section{Performance Evaluation}\label{sec:performance}
\subsection{Implementation Details}
We configure the simulation with \(
	{\numresourcesavailable = \num{16}}
\)~available resources and \(
	{\numusers = \num{10}}
\) users.
User profiles are set up according to \reftab{tab:user_profiles}, with five~'Normal', two~'High Datarate', two~'Low Latency' users, and one~'Emergency Vehicle'.
For the given configuration, a job creation probability~\(
	{\probabilityjobcreation = \SI{20}{\percent}}
\) for each simulation step and user puts a high expected load of \( \num{1.6} \)~requests per available resource on average on the schedulers.
We run a total of \( \num{10000} \)~episodes at \( \num{50} \)~time steps~\( \timeindex \) per episode for training and evaluation each, with the mini batch size set to~\( {\batchsize = 64} \) sampled experiences per training step.
Parameter optimization via \stochasticgradientdescent is carried out by the Adam optimizer~\cite{kingma_adam_2015} with default settings and a learning rate of~\( \num{1e-4} \).
Reward weightings are set to \( {\weightrewardcapacity = \weightrewardpacketrate = \num{0.25}} \), \( {\weightrewardtimeout = \weightrewardtimeoutev = \num{1.0}} \), giving roughly equal significance to each goal metric with the average expected magnitude of the respective rewards.

\begin{table}[tb]
	\renewcommand{\arraystretch}{1.3}
	\caption{User Profiles}
	\label{tab:user_profiles}
	\begin{center}
		\begin{tabular}{c|c|c}
			\hline
			& \textbf{Delay} & \textbf{Max Job Size} \( \resourceblock_{\text{\text{max}}} \) \\ 
			& in sim. steps		& in res. blocks		\\\hline\hline
			Normal				& \( \num{20} \)				& \( \num{30} \) \\
			High Packet Rate	& \( \num{20} \)				& \( \num{40} \) \\
			Low Latency			& \( \num{2} \)					& \( \num{8} \) \\
			Emergency Vehicle	& \( \num{1} \)					& \( \num{16} \) \\
			\hline
		\end{tabular}
	\end{center}
\end{table}

For the algorithm selection \dqn, a five layer feed-forward network is selected.
Layers have \( \num{300} \)~nodes each, except for the last layer that was split in two branches with \( \num{200} \)~nodes each according to the DuelingDQN structure~\cite{hessel2018rainbow}.
Exploration is done by selecting random actions at an initial chance of~\(
	{\explorationchance = \SI{99}{\percent}}
\), decaying linearly to~\( \SI{0}{\percent} \) after~\( \SI{80}{\percent} \) of training episodes.
The exponential future reward decay factor is set to~\( {\rewarddiscount = \num{0.9}} \) in order to put significance on only a low number of future steps. User TX-SNR~\( {\signalpower / \noisepower} \) is fixed to \( \SI{13}{\dB} \).

We implement the simulation in python using the tensorflow library primarily. The full implementation is made available in~\cite{gracla2020code}.

\subsection{Results}

As the simulation model contains stochastic components, the results achievable on each metric have an inherent variance depending on the specific realizations.
For example, a spike in generated jobs will result in timeouts irrespective of the scheduling method.
For this reason, results achieved during testing are displayed in cumulative histograms.

\begin{figure*}[tb]
	\centering
	\subfloat[]{
		\input{figures/scheduling_testing_capacities.pgf}
		\label{fig:results_capacity}
	}
	\hfil
	\subfloat[]{
		\input{figures/scheduling_testing_datarate_satisfaction.pgf}
		\label{fig:results_packet_rate}
	}
	\\
	\subfloat[]{
		\input{figures/scheduling_testing_latency_violations.pgf}
		\label{fig:results_latency_violations}
	}
	\hfil
	\subfloat[]{
		\input{figures/scheduling_testing_timeouts_ev_only.pgf}
		\label{fig:results_ev_latency_violations}
	}
	\caption{%
		Cumulative histograms of the schedulers' performance on the individual parts (a) \( \rewardcapacity \), (b) \( \rewardpacketrate \), (c) \( \rewardtimeout \) and (d) \( \rewardtimeoutev \) of the overall optimization goal \( \reward \) (compare \refeq{eq:reward}). Dominating all submetrics at once is impossible as the individual submetrics have conflicting objectives, e.g., scheduling an urgent job while the channel is weak. Maximum Throughput~(MT), Max-Min-Fair~(MMF) and Delay Sensitive~(DS) schedulers show strengths in some submetrics but weaknesses in others. The presented DQN scheduler achieves its overall optimization goal by balancing strong performance in all submetrics.
	}
	\label{fig:results_submetrics}
\end{figure*}

\begin{figure}[tb]
	\centering
	\input{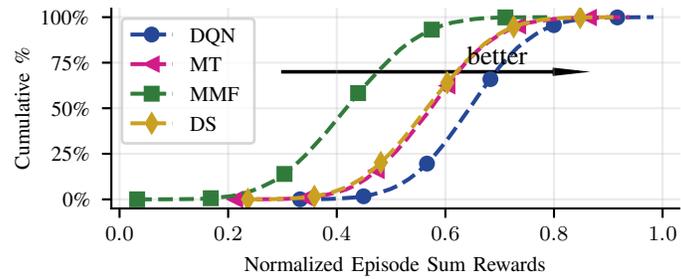}
	\caption{Cumulative histogram of performance of Maximum Throughput~(MT), Max-Min-Fair~(MMF) and Delay Sensitive~(DS) schedulers as well as DQN adaptive scheduler on the weighted sum reward metric. For each episode, all achieved rewards \( \reward \) are summed. Achieved reward sums are normalized by the highest achieved reward sum.}
	\label{fig:results_reward}
\end{figure}

\reffig{fig:results_reward} shows each schedulers performance on the combined reward metric~\( \reward \).
On this metric, the \dqn adaptive scheduler is able to find a strategy that outperforms any single model-based algorithm.
Breaking the sum metric~\( \reward \) down into its constituent parts sheds light on how this is achieved. As shown in \reffig{fig:results_submetrics}, the \dqn scheduler balances the submetrics against each other, yielding some performance on each of them compared to the benchmark.

\begin{figure}[tb]
	\centering
	\input{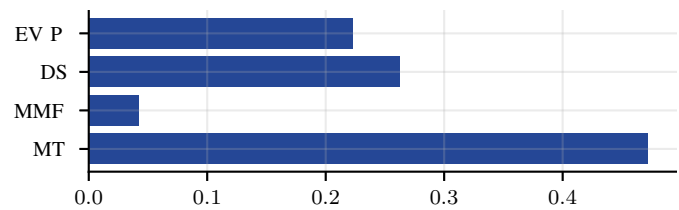}
	\caption{Relative selection rate of EV Priority~(EV P), Delay Sensitive~(DS), Max-Min-Fair~(MMF) and Maximum Throughput~(MT) schedulers during testing.}
	\label{fig:results_algorithm_selections}
\end{figure}

Of particular interest are the \emergencyvehicletext-specific timeouts in \reffig{fig:results_ev_latency_violations}.
This metric is not specifically targeted by any of the model-based algorithms depicted.
A modest double-weighting of \emergencyvehicletext specific timeouts within the goal metric~\( \reward \), combined with the introduction of the otherwise sub-optimal \emergencyvehicletext Priority scheduling algorithm to the selection pool, has enabled the \dqn based scheduler to significantly suppress timeouts in Emergency Vehicles even compared to the otherwise timeout focused Delay Sensitive algorithm without hurting overall performance overmuch.
As \reffig{fig:results_algorithm_selections} shows, the \emergencyvehicletext Priority algorithm was only selected a comparably low amount of times to achieve this goal.

The \dqn{}s learned behavior can also be monitored to reveal underlying features of the simulation.
As \reffig{fig:results_algorithm_selections} highlights, for the given reward weighting, the \maxminfair algorithm was only selected a low number of times.
Further investigation could reveal whether these are, for example, high impact outlier cases that could be better served with an additional model-based algorithm that is specifically targeted to them, or whether the \maxminfair algorithm is not fit for the given scenario.

The \dqn decision making does however add another layer of computation to the scheduling process.
Further, while the ensemble method can relax the sophistication required from each model-based part of the ensemble, the composite scheduling function can only assume the function space spanned by the group of model-based algorithms.
In other words, the \dqn adaptive method is unable to discover strategies that go beyond combining the available models.
Determining whether the selection of model-based algorithms provided is rich enough to serve the problem therefore remains a burden on the designer.

\section{Conclusion}\label{sec:conclusion}
In this paper we presented a \reinforcementlearning-based communications resource scheduler that constructs an effective scheduling paradigm from an ensemble of model-based algorithms.
We achieved this by learning to switch to whichever algorithm promises the highest expected long term benefit based on the current queue state.
This approach combines the flexible goal optimization of \reinforcementlearning methods with the rigid predictability of model-based algorithms.
It is noteworthy for applications with complex, conflicting performance goals, where either a)~strong models exist to cover parts of the problem, or b)~explicit modeling is otherwise necessary, e.g., due to very low tolerance for outliers, such as the transmission of critical medical data.
Using this approach unlocks the benefits of \reinforcementlearning without abandoning explicit modeling.
For the simulation presented, the adaptive model-switching scheduler was able to learn to outperform single, model-based algorithms on a weighted sum utility metric.

%
%
%
\bibliographystyle{ref/IEEEtran}
{%
	\makeatletter  
	\clubpenalty10000  
	\@clubpenalty \clubpenalty
	\widowpenalty10000
	\makeatother
	
	\bibliography{ref/IEEEabrv,ref/references}
}%

\end{document}